\documentclass[10pt,twocolumn,letterpaper]{article}

\usepackage{cvpr}              
\usepackage{graphicx}
\usepackage{amsmath}
\usepackage{amssymb}
\usepackage{booktabs}
\usepackage{colortbl}
\usepackage[pagebackref,breaklinks,colorlinks]{hyperref}

\definecolor{cvprblue}{rgb}{0.21,0.49,0.74}

\def\paperID{8105} 
\def\confName{CVPR}
\def\confYear{2025}

\title{MamKPD: A Simple Mamba Baseline for Real-Time 2D Keypoint Detection}

\author{
Yonghao Dang \textsuperscript{1,}$^{\dagger}$ \quad
Liyuan Liu\textsuperscript{1,}$^{\dagger}$ \quad
Hui Kang\textsuperscript{2} \quad
Ping Ye \textsuperscript{1} \quad
Jianqin Yin\textsuperscript{1,}$^*$ \quad \\
\textsuperscript{1} Beijing University of Posts and Telecommunications \\
\textsuperscript{2} Inspur Genersoft Co., Ltd. \\
{\tt\small \{dyh2018, liuliyuan2023, jqyin\}@bupt.edu.cn}
}

\begin{document}
\maketitle
\let\thefootnote\relax\footnotetext{
$^\dagger$Equal contribution.}
\let\thefootnote\relax\footnotetext{
$^*$Corresponding authors.}

\maketitle
\begin{abstract}
Real-time 2D keypoint detection plays an essential role in computer vision. Although CNN-based and Transformer-based methods have achieved breakthrough progress, they often fail to deliver superior performance and real-time speed. This paper introduces MamKPD, the first efficient yet effective mamba-based pose estimation framework for 2D keypoint detection. The conventional Mamba module exhibits limited information interaction between patches. To address this, we propose a lightweight contextual modeling module (CMM) that uses depth-wise convolutions to model inter-patch dependencies and linear layers to distill the pose cues within each patch. Subsequently, by combining Mamba for global modeling across all patches, MamKPD effectively extracts instances' pose information. We conduct extensive experiments on human and animal pose estimation datasets to validate the effectiveness of MamKPD. Our MamKPD-L achieves \textbf{77.3\% AP} on the COCO dataset with \textbf{1492 FPS on an NVIDIA GTX 4090 GPU.}  Moreover, MamKPD achieves state-of-the-art results on the MPII dataset and competitive results on the AP-10K dataset while \textbf{saving 85\% of the parameters} compared to ViTPose. Our project page is available at \href{https://mamkpd.github.io/}{https://mamkpd.github.io/}.
\end{abstract}    
\section{Introduction}
\label{sec:intro}

Real-time 2D keypoint detection, which requires the model to locate interesting points of instances with low latency, has a wide range of applications in many fields because it can provide pure action information for downstream tasks, such as virtual reality \cite{DHRNet}, action assessment \cite{RPSTN}, and human-robot interaction \cite{VR}. Despite the breakthrough advancements in 2D keypoint detection, exemplified by CNN-based \cite{AlphaPose,Hourglass,Openpose,SBL,HRNet,RPSTN} and Transformer-based keypoint detection frameworks \cite{TokenPose,TFPose,PRTR,PETR,PPT,SwinPose,OTPose,GroupPose}, these methods are often constrained by network scale, typically requiring expensive computing resources.

\begin{figure}
    \centering
    \includegraphics[scale=0.1]{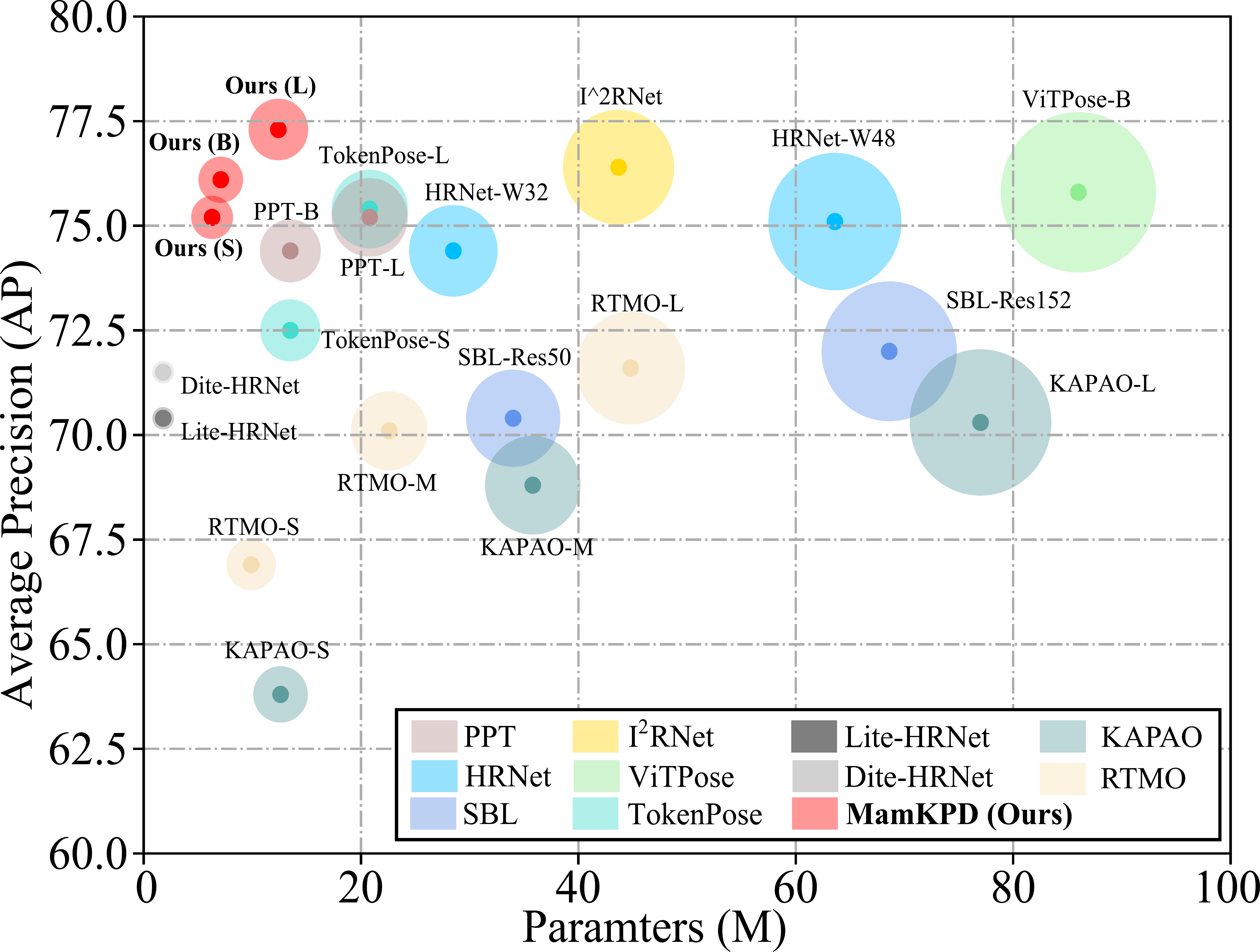}
    \caption{Comparisons of performance and parameters on COCO val2017 set. The circles' size represents the model's scale, \textit{i.e.}, the number of the model's parameters.}
    \label{fig:fig1}
\end{figure}

Numerous studies have sought to enhance the efficiency of keypoint detection by compressing network size or adopting lightweight network architectures. Network compression approaches \cite{Lite-Pose,Lite-HRNet,Dite-HRNet,PPT} generally employ pruning or replace components with more efficient alternatives to increase inference speed. Lightweight network methods \cite{EfficientPose,EfficientHRNet,KAPAO,RTMO,RTMPose}, on the other hand, use more efficient backbone networks, such as YOLO \cite{YOLO}, EfficientNet \cite{EfficientNet}, to improve inference efficiency. However, most methods achieve increased efficiency at the expense of detection precision. As shown in Figure \ref{fig:fig1}, although some approaches reduce the model's parameters, they also experience a significant drop in performance. This raises a challenging question: \textit{how can we improve model efficiency without compromising accuracy?} 

Recently, state space models (SSMs) \cite{SSM}, exemplified by Mamba \cite{Mamba}, have demonstrated superior sequential modeling capabilities while maintaining high efficiency. Inspired by the successful application of Mamba in computer vision tasks \cite{Vim,Vmamba,Swin-UMamba,li2024videomamba,PoseMamba,MambaSurvey}, this study is the first to explore the potential of Mamba for 2D keypoint detection. Nevertheless, \textit{the traditional mamba block merely aggregates image patches during updating states, which limits its capacity to capture contextual features (inter-patch dependencies)}. Contextual information of target instances is crucial for the keypoint detection, as the relationships between patches allow the model to learn structural information of the instance.

To address the above issues, we propose a novel 2D keypoint detection framework called MamKPD, marking the first \textbf{Mam}ba-based baseline network for 2D \textbf{K}ey\textbf{P}oint \textbf{D}etection, as illustrated in Figure \ref{fig:overview}. Additionally, we introduce a lightweight contextual modeling module (CMM) designed to enhance conmmunication between patches. Specifically, MamKPD consists of a stem, a Mamba-based encoder, and a simple keypoint decoder. In MamKPD, we first use a simple stem to extract primary pose features from the input, which helps filter out irrelevant noise. Next, a three-stage Mamba-based encoder is employed to extract pose features further. At each stage, a contextual modeling module is used to capture dependencies between patches, followed by the 2D-selective scan (SS2D) module to model the interactions across all patches. Finally, a keypoint decoder generates the keypoint heatmaps. Thanks to its streamlined architecture, MamKPD effectively balances network efficiency and accuracy. We validate the effectiveness of MamKPD on both human and animal pose estimation datasets. Experimental results demonstrate that MamKPD achieves competitive precision while maintaining high efficiency.

The contributions of this paper are summarized below.
\begin{itemize}
    \item We propose a novel and efficient 2D keypoint detection network named MamKPD, marking the first Mamba-based architecture designed specifically for keypoint detection tasks. The proposed MamKPD surpasses existing approaches in inference speed while maintaining competitive keypoint localization accuracy.
    \item We introduce a simple contextual modeling module (CMM) to enhance the original Mamba module, thereby improving its capacity for contextual feature modeling. The CMM module, in conjunction with the SS2D module, extracts rich pose-related information by capturing inter- and intra-patch features.
    \item Without bells and whistles, MamKPD achieves promising results across human and animal 2d keypoint detection datasets, including the COCO \cite{COCO}, MPII \cite{MPII},and AP-10K \cite{AP-10K} datasets. On a single Nvidia 4090 GPU, MamKPD-L achieved an inference speed of 1492 FPS with a keypoint detection accuracy of 77.3\%, surpassing existing real-time methods. 
\end{itemize}

\section{Related Works}
\label{sec:related_works}

\subsection{General 2D Keypoint Detection}
2D keypoint detection has made significant breakthroughs in recent years. Among the mainstream methods, CNN-based and Transformer-based detection architectures are particularly representative.

\textbf{CNNs-based approaches.} Convolutional Neural Networks (CNNs) were the dominant backbone network for early 2D keypoint detection methods. CNN-based 2D keypoint detection approaches can be roughly divided into two categories: \textit{regression-based} \cite{Deeppose, IEF, sun2017compositional, luvizon2019human} and \textit{heatmap-based} ~\cite{jain2013learning,tompson2014joint,chen2014articulated,tompson2015efficient,hu2016bottom,chu2017multi,yang2017learning,SBL,HRNet}. \textit{Regression-based methods}, such as DeepPose \cite{Deeppose} and IEF \cite{IEF}, aim to directly obtain the coordinates $(x, y)$ of each keypoint from the image. Unlike regression-based methods, \textit{heatmap-based approaches} regress each key point’s heatmap from the image rather than the coordinate points. On the one hand, this preserves the structural information of the image; on the other hand, it facilitates network training. Currently, heatmap-based methods, such as HRNet \cite{HRNet} and SBL \cite{SBL}, are still widely used. In this paper, to ensure the keypoint localization performance of MamKPD, we design MamKPD as a heatmap-based framework. 


\textbf{Transformer-based approaches.} Transformer architectures have been widely used in 2D keypoint detection tasks \cite{TFPose, PRTR,mao2022poseur,TokenPose,CLAMP}. TFPose \cite{TFPose} first introduced a Transformer to the pose estimation task. PRTR \cite{PRTR} employed cascading Transformers and treated the pose recognition task as a regression task. To explore the potential of a transformer for human pose estimation, ViTPose \cite{ViTPose} employed plain and non-hierarchical vision transformers as backbones to improve accuracy, achieving state-of-the-art performance. Additionally, some methods exploit Transformers' superior feature fusion capabilities, using multi-modal data, such as text and image, to enhance network performance \cite{CLAMP,LAMP}. Despite the impressive performance of Transformer-based methods, the quadratic computational complexity of the attention mechanism requires significant resources, limiting their widespread application.

\subsection{Real-Time 2D Keypoint Detection}
Real-time 2D keypoint detection tasks have gained widespread attention recently, as real-time performance is crucial for practical applications. Most existing real-time keypoint detection methods can be broadly categorized: \textit{1)} Pruning-based methods, which improve model efficiency by removing redundant layers \cite{Lite-HRNet,Lite-Pose,Dite-HRNet,EfficientHRNet} or task-irrelevant tokens \cite{PPT,GTPT} and integrating high-performance computation modules; \textit{2)} Lightweight backbone-based methods \cite{EfficientNet,EfficientPose,KAPAO,RTMPose,RTMO}, which reduce model latency by using more efficient backbone networks, \textit{e.g.} EfficientNet \cite{EfficientNet}, YOLO \cite{YOLO} and so on, to extract pose features. Representative real-time keypoint detection methods, such as RTMPose \cite{RTMPose} and RTMO \cite{RTMO}, have achieved impressive detection accuracy and efficiency. 

Despite the significant progress made by the above methods, they still struggle to balance accuracy and efficiency. Most approaches continue to sacrifice network accuracy to improve model inference speed. To address this issue, this paper aims to design a simple and efficient keypoint detection network that significantly enhances model inference speed while maintaining high accuracy.

\subsection{State Space Model}
Recently, spatial state models \cite{fu2022hungry,gu2020hippo,gu2021efficiently}, exemplified by Mamba \cite{Mamba}, have demonstrated superior performance in natural language processing (NLP). Mamba \cite{Mamba} has achieved impressive results through its state selection mechanism and hardware control design, making it a strong contender for attention-based transformer architectures. Inspired by advancements in the NLP, researchers have successfully extended Mamba to computer vision tasks~\cite{Vmamba,Swin-UMamba,PoseMamba, huang2024localmamba, yang2024plainmamba, li2024videomamba, yang2024remamber}. Similarly, Mamba has demonstrated superior performance and inference efficiency in computer vision tasks. However, the exploration of Mamba's potential in 2D keypoint detection remains untapped. In this paper, we do not simply apply SSM to pose estimation. We explore generic structures of Mamba for the task of 2D keypoint detection. Although Mamba has shown superior performance in various computer vision tasks, no research has explored its potential in 2D keypoint detection. This paper first investigates Mamba's capabilities for 2D keypoint detection.
\section{Methodology}
\label{sec:method}
\begin{figure*}[h]
    \centering
    \includegraphics[scale=.44]{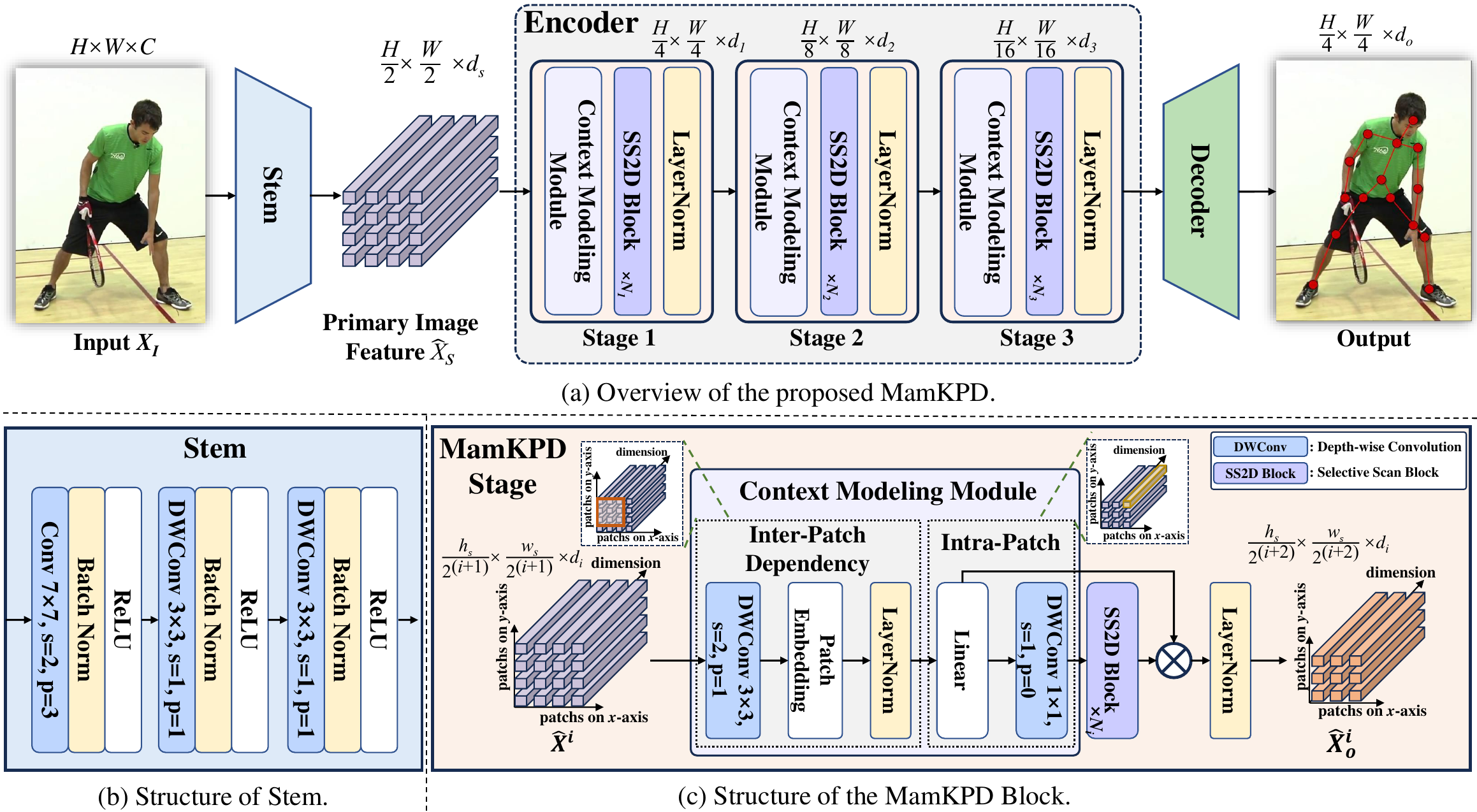}
    \caption{Overview of the proposed MamKPD is illustrated in Figure (a). Figures (b) and (c) display Stem and MamKPD stage structures.}
    \label{fig:overview}
\end{figure*}

The proposed MamKPD, implemented using the top-down paradigm, is a simple yet effective pose estimation framework. Following the SBL proposed by \cite{SBL}, MamKPD also adopts an encoder-decoder structure to extract refined pose features and decode pure joint heatmaps. Next, we introduce the key elements of MamKPD.

\subsection{Overview}
The overview of MamKPD is illustrated in Figure \ref{fig:overview}. MamKPD consists of a CNN-based stem network, a Mamba-based encoder, and a simple decoder. Given an input image $X_I \in \mathbb{R}^{H \times W \times3}$, where $H$ and $W$ are the height and width of the input image. MamKPD aims to regress $K$ joint heatmaps (\textit{i.e.}, $\mathbf{H} \in \mathbb{R}^{h \times w \times K}$, where $h$ and $w$ refer as the height and width of the joint heatmaps) for each person in the image.

\textbf{Stem.} Stem filters out irrelevant information, such as background, from the image and extracts the initial pose features. Specifically, stem takes the image $X_I$ as input and extracts primary pose features $\hat{X}_{S}$, which serve as inputs for the following mamba encoder. As shown in Figure \ref{fig:overview}(b), we utilize a sizeable convolutional layer with a kernel size of 7, along with two depth-wise convolutions with a kernel size of 3, to preliminarily filter out redundant information from the original image.  

\begin{equation}
    \hat{X}_{S} = Stem\left(X_I\right), \hat{X}_{S} \in \mathbb{R}^{\frac{H}{2} \times \frac{W}{2} \times d_s}
    \label{eq:stem}
\end{equation}
where $\hat{X}_{S}$ represents the preliminary features extracted by the stem,  which capture low-level features such as the instance's contour while filtering out background, color, and other irrelevant information. $d_s$ is the feature dimension.

\textbf{Mamba-based encoder.} The encoder consists of three stages, each of which is composed of a context modeling module (CMM), $N_i$ SS2D blocks (where $ i \in \{1,2,3\}$ represents the $i$-th stage), and a layer normalization.

MamKPD encoder takes the primary image feature $\hat{X}_s$ as input and outputs the extracted multi-level pose features, which serve as inputs for the following simple decoder. In the MamKPD encoder, we use 3 feature levels for three stages, with downsampling rates of \{4, 8, 16\}. The processing flow of the encoder can be expressed as follows.
\begin{equation}
    \hat X_o^i = LN\left( {SS2D\left( {CMM\left( {{{\hat X}^i}} \right)} \right)} \right)
    \label{eq:encoder}
\end{equation}
where $\hat X_o^i$ is the output of the $i$-th stage. $LN\left( \cdot \right)$ and $SS2D\left( \cdot \right)$ represent the layer normalization and visual state space block. $CMM\left( \cdot \right)$ is the proposed context modeling module. Since Mamba extracts image features patch-by-patch, the information communication between patches is limited. Here, we design a context modeling module(introduced in section \ref{subsec:MamKPD}) for learning inter-patch dependency and enhancing the multi-scale feature extraction capability of the MamKPD encoder. 

\textbf{Decoder.} We directly follow SBL \cite{SBL} and ViTPose \cite{ViTPose} to build a simple decoder to transfer each person's pose feature $\hat{X}_o$ into $K$ joint heatmaps, $\mathbf{H} \in \mathbb{R}^{h \times w \times K}$. Specifically, we employ two deconvolutional layers with a kernel size of 4 as the decoder to upsample the features extracted by the encoder and decode them into keypoint heatmaps.
\begin{equation}
    \mathbf{H} = Decoder\left(\hat{X}_o\right), \mathbf{H} \in \mathbb{R}^{\frac{H}{4} \times \frac{W}{4} \times K}
\end{equation}

\textbf{Loss function.} In this paper, we try to simplify the training process of MamKPD. Therefore, we only adopt the commonly used mean squared error (MSE) loss \cite{SBL,HRNet} to train the proposed MamKPD, as described in Equation \ref{eq:loss}. Without any extra supervision, the proposed MamKPD achieves promising performance on 2D keypoint detection.
\begin{equation}
    Loss = \frac{1}{M}\sum_{i=0}^M\left(\mathbf{H}_{i} - \Bar{\mathbf{H}}_{i}\right)^2
    \label{eq:loss}
\end{equation}
where $M$ is the total number of training samples. $\mathbf{H}_i$ and $\Bar{\mathbf{H}}_i$ denote the $i$-th person's joint heatmaps predicted by the model and the ground truth.

\subsection{Structure of MamKPD Stage}
\label{subsec:MamKPD}
The structure of each proposed MamKPD stage consists of a context modeling module (CMM) and $N_i$ SS2D blocks (where $N_i$ represents the number of SS2D blocks in the $i$-th stage), as shown in Figure \ref{fig:overview}(c).

\textbf{Context modeling module.} The traditional Mamba module recursively extracts features from patches while maintaining constant feature dimensions during the modeling process, leading to two inherent limitations of Mamba. \emph{First, Mamba struggles to model long-range dependencies between distant patches that are not adjacent. Second, its ability to extract multi-scale features is limited.} Therefore, we propose a context modeling module (CMM) to assist Mamba in extracting multi-scale image features and enhance the information communication between different patches. 

As shown in Figure \ref{fig:overview}(c), to balance the model's scale and performance, CMM is designed as a simple module comprising two stages: inter-patch dependency modeling and intra-patch feature extraction. For \textit{inter-patch dependency modeling}, there is a depth-wise convolution (DWConv), a patch embedding layer, and a layer normalization. Specifically, the \emph{DWConv} captures dependencies between patches within the receptive field by performing convolution operations along the patch dimension while downsampling the input features to provide a larger receptive field for subsequent modules. The \emph{patch embedding layer}, which includes patch partitioning and encoding, further extracts inter-patches features through the additional DWConv. Notably, since the previous DWConv performs downsampling on the patches, the input resolution for the patch embedding layer is reduced, but the receptive field is expanded. This allows the patch embedding to model dependencies between more distant patches. \textit{Intra-patch feature extraction} aims to capture pose-relevant cues from each image patch. We use a linear layer and a $1 \times 1$ depth-wise convolution to extract features from each patch. The CMM modeling process can be formalized as follows.

\begin{equation}
    \begin{array}{cc}
    & \hat{X}_{inter} = LN\left(Embed\left(DWConv\left( \hat{X}^{i} \right)\right)\right) \\
    & \hat{X}_{CMM} = DWConv\left(Linear\left(\hat{X}_{inter}\right)\right)
    \end{array}
\end{equation}
where $\hat{X}_{CMM} \in \mathbb{R}^{\frac{H}{2^{(i+1)}} \times \frac{W}{2^{(i+1)}} \times d_i},i\in \{1,2,3\}$, represents the contextual feature learned by CMM.

\textbf{SS2D block.} The SS2D module aggregates information from all image patches, enabling the model to learn a complete representation of the pose structure. The contextual feature $\hat{X}_{CMM}$ extracted by CMM is passed through the SS2D block to integrate patch information. Subsequently, the features aggregated by SS2D are used as attention weights, fused with the previous intra-patch features to activate the most important image patches. 

Furthermore, the patch scanning strategy is crucial for the SS2D module. Following \cite{Vmamba}, this paper employs a sequential scanning strategy in four directions to rearrange patches. Since the CMM module models inter-patch dependencies to some extent, a simple sequential patch reordering is used here to further capture information between adjacent patches.

\section{Experiment}
\label{sec:expe}

\subsection{Settings}
\textbf{Datasets.} We use human pose estimation datasets, such as COCO~\cite{COCO} and MPII~\cite{MPII}, along with animal pose estimation datasets like AP-10K~\cite{AP-10K}, to validate the potential of MamKPD in 2D keypoint detection tasks. 

\emph{COCO dataset} \cite{COCO} contains over 200,000 images and 250,000 person instances labeled with 17 key points. Following \cite{RTMO,ViTPose}, we train MamKPD on the COCO train2017 set and evaluate it on the val2017 set.

\emph{MPII dataset} \cite{MPII} includes around 25k images containing over 40k people with 16 annotated body joints. Following existing methods \cite{TokenPose,ViTPose}, MamKPD is trained on the MPII training set and evaluated on the validation set. 

\emph{AP-10k dataset} \cite{AP-10K} contains 10,015 images collected and filtered from 23 animal families and 54 species, the largest and most diverse dataset for animal pose estimation. Following \cite{AP-10K,ViTPose}, we train and validate MamKPD on the AP-10K training and test sets.

\textbf{Implementation details.} We provide three variations of the MamKPD structure: small, base, and large models, as shown in Table \ref{tab:architecture}. The small model does not include the Stem module, while the base and large models use different feature dimensions. MamKPD follows the top-down paradigm. In this paradigm, the single-person instance is first detected by an instance detector \cite{HRNet,ViTPose}, and then key points are predicted. We use the Adam \cite{Adam} optimizer during training with an initial learning rate of 1e-3. The learning rate is reduced by 10 at the 170 and 200 epochs. The total training process requires 210 epochs. During training, we adopt the common data augmentations \cite{HRNet,SBL,TokenPose,PPT}, including random flip, crop, and resize. All experiments are conducted on two RTX 4090 GPUs.

\begin{table}[htbp]
    \centering
    \small
    \setlength\tabcolsep{1.5pt}
    \caption{Model specifications of MamKPD variants. \textit{Blocks} represent the number of mamba blocks in each stage. \textit{Dimension} denotes the number of channels of features in different stages.}
    \begin{tabular}{l|c|cc|cc}
    \toprule
    \rowcolor{gray!30} Model & Stem & Blocks & Dimension & \#Param. & GFLOPs \\
    \hline
        MamKPD-Small & $\times$ & [2,4,6] & [96,192,384] & 6.3M & 0.5 \\
        MamKPD-Base & $\checkmark$ & [2,4,6] & [96,192,384] & 7.1M & 3.0 \\
        MamKPD-Large & $\checkmark$ & [2,4,6] & [128,256,512]& 12.4M & 4.3 \\
    \bottomrule
    \end{tabular}
    \label{tab:architecture}
\end{table}

\textbf{Evaluation metric.} For the COCO and AP-10K datasets, we evaluate performance using accuracy at various thresholds and target sizes, including Average Precision ($AP$), AP at 50\% ($AP_{50}$), and AP at 75\% ($AP_{75}$), as well as Average Precision for medium ($AP_{M}$) and large ($AP_L$) targets. For the MPII dataset, we use the Percentage of Correct points (PCK) as the evaluation metric for each key point, with the mean PCK serving as an overall performance metric.

\subsection{Comparisons with the State-of-the-arts}

We aim to build a simple baseline based on mamba for 2D keypoint detection. Thus, we mainly compare our MamKPD with the previous backbones designed for pose estimation. This section evaluates the proposed MamKPD from two aspects: keypoint localization accuracy and inference speed. 

\subsubsection{Analysis of Keypoint Localization Accuracy}

\textbf{Comparisons with existing methods on COCO dataset.} Table \ref{tab:coco} presents the comparisons on the COCO val2017 set. On COCO val2017, MamKPD surpasses the Lite-HRNet and Dite-HRNet. MamKPD-S achieves competitive performance with PPT while saving 53.3\% in parameter count. Furthermore, MamKPD-L outperforms PPT-L by 2.1\% AP, with approximately 40\% fewer parameters. Although MamKPD-L scores 1\% AP lower than ViTPose, it requires only 4\% of ViTPose's parameters. Moreover, MamKPD-B achieves performance exceeding that of ViTPose-B with only 8.3\% of ViTPose-B's parameter count. These results demonstrate the effectiveness of MamKPD.

\begin{table*}[htbp]
    \centering
    \small
    \setlength\tabcolsep{4.3pt}
    \caption{Comparisons with state-of-the-art methods on the COCO val2017 set \cite{COCO}. \textbf{\textit{Bold}} text indicates the \textbf{\textit{best results}}, while \underline{\textit{underlined}} text denotes the \underline{\textit{second-best results}}. $\dagger$ denotes the methods reproduced on our server.}
    \begin{tabular}{l|c|c|ccc|cccccc}
    \toprule
    \rowcolor{gray!30} Method & Backbone & Pretrain & \#Params & GFLOPs & Speed (fps) & $AP$ & $AP_{50}$ & $AP_{75}$ & $AP_{M}$    & $AP_{L}$ & AR  \\
    \hline
    \multicolumn{11}{c}{Conventional Methods} \\
    \hline
    SBL-Res50 \cite{SBL} & ResNet-50 & Y & 34M & 8.9 & - & 70.4 & 88.6 & 78.3 & 67.1 & 77.2 & 76.3 \\
    SBL-Res152 \cite{SBL} & ResNet-152 & Y & 68.6M & 15.7 & 829 & 72.0 & 89.3 & 79.8 & 68.7 & 78.9 & 77.8  \\
    HRNet-W32 \cite{HRNet} & HRNet-W32 & Y & 28.5M & 7.1 & 916 & 74.4 & 90.5 & 81.9 & 70.8 & 81.0 & 79.8  \\
    HRNet-W48 \cite{HRNet} & HRNet-W48 & Y & 63.6M & 14.6 & 649 & 75.1 & 90.6 & 82.2 & 71.5 & 81.8 & 80.4 \\
    TokenPose-S \cite{TokenPose} & HRNet-W48 & Y & 13.5M & 2.2 & 602 & 72.5 & 89.3 & 79.7 & 68.8 & 79.6 & 78.0  \\
    TokenPose-L \cite{TokenPose} & HRNet-W4 & Y & 20.8M & 9.1 & 309 & 75.4 & 90.0 & 81.8 & 71.8 & 82.4 & 80.4  \\
    GroupPose-L \cite{GroupPose} & Swin-L & Y & - & -  & - & 74.8 & \underline{91.6} & 82.1 & 69.4 & 83.0 & - \\
    $I^2$R-Net \cite{I2RNet} & HRFormer-B & Y & 43.7M & 12.8  & - & 76.4 & 90.8 & 83.2 & 72.3 & \underline{83.7} & 81.4  \\
    ViTPose-B \cite{ViTPose} & Transformer & Y & 86.0M & 17.1  & 944 & 75.8 & 90.7 & 83.2 & 68.7 & 78.4 & 81.1  \\
    ViTPose-L \cite{ViTPose} & Transformer & Y & 307.9M & -   & 411 & \textbf{78.3} & 91.4 & \textbf{85.2} & 71.0 & 81.1 & \underline{83.5} \\
    SwinPose \cite{SwinPose} & Swin-L & Y & 196.4M & 202.6 & - & 76.3 & \textbf{93.5} & \underline{83.4} & \underline{72.5} & 81.7 & \textbf{84.7} \\
    \hline
    \multicolumn{11}{c}{Lightweight Methods} \\
    \hline
    Lite-HRNet \cite{Lite-HRNet} & Lite-HRNet-30 & N & \textbf{1.8M} & \underline{0.7} & - & 70.4 & 88.7 & 77.7 & 67.5 & 76.3 & 76.2  \\
    Dite-HRNet \cite{Dite-HRNet} & Dite-HRNet-30 & N & \textbf{1.8M} & \underline{0.7} & - & 71.5 & 88.9 & 78.2 & 68.2 & 77.7 & 77.2  \\
    PPT-B \cite{PPT} & Transformer & Y & 13.5M & 5.0 & - & 74.4 & 89.6 & 80.9 & 70.8 & 81.4 & 79.6  \\
    PPT-L \cite{PPT} & Transformer & Y & 20.8M & 8.7 & - & 75.2 & 89.8 & 81.7 & 71.7 & 82.1 & 80.4  \\
    \hline
    \multicolumn{11}{c}{Real-Time Methods} \\
    \hline
    KAPAO-$\rm S^{\dagger}$ \cite{KAPAO} & CSPNet & Y  & 12.6M & - & 1037 & 63.8 & 88.4 & 70.4 & 58.6 & 71.7 & 71.2  \\
    KAPAO-$\rm M^{\dagger}$ \cite{KAPAO} & CSPNet & Y & 35.8M & - & 972 & 68.8 & 90.5 & 76.5 & 64.3 & 76.0 & 76.3  \\
    KAPAO-$\rm L^{\dagger}$ \cite{KAPAO} & CSPNet & Y & 77.0M & -  & 952 & 70.3 & 91.2 & 77.8 & 66.3 & 76.8 & 77.7 \\
    RTMO-$\rm S^{\dagger}$ \cite{RTMO} & CSPDarknet & Y & 9.9M & - & 1522 & 66.9 & 88.8 & 73.6 & 61.1 & 75.7 & 70.9  \\
    RTMO-$\rm M^{\dagger}$ \cite{RTMO} & CSPDarknet & Y & 22.6M & - & 1423 & 70.1 & 90.6 & 77.1 & 65.1 & 78.1 & 74.2  \\
    RTMO-$\rm L^{\dagger}$ \cite{RTMO} & CSPDarknet & Y & 44.8M & -  & 1282 & 71.6 & 91.1 & 79.0 & 66.8 & 79.1 & 75.6 \\
   
    \hline
    \textbf{MamKPD-S (Ours)} & Mamba & N & \underline{6.3M} & \textbf{0.5} & \textbf{2564} & 75.2 & 90.4 & 82.2 & 71.5 & 81.9 & 75.3  \\
    \textbf{MamKPD-B (Ours)} & Mamba & N & 7.1M & 3.1 & \underline{2439} & 76.1 & 90.6 & 82.6 & 72.4 & 82.9 & 80.9  \\
    \textbf{MamKPD-L (Ours)} & Mamba & N & 12.4M & 4.3 & 1492 & \underline{77.3} & 90.8 & \underline{83.4} & \textbf{73.6} & \textbf{84.0} & 82.1  \\
    \bottomrule
    \end{tabular}
    \label{tab:coco}
\end{table*}

\textbf{Comparisons with existing methods on MPII dataset.} Table~\ref{tab:MPII} presents a comparison of MamKPD with existing methods on the single-person pose estimation dataset MPII \cite{MPII}. As shown in Table \ref{tab:MPII}, the overall performance of MamKPD-S already surpasses existing methods. Additionally, MamKPD-L achieves the best localization results for each keypoint, with notable improvements in wrist and ankle localization accuracy compared to prior methods.

\begin{table}[htbp]
    \small
    \centering
    \setlength\tabcolsep{1.4pt}
    \caption{Comparisons with state-of-the-art methods on the MPII validation set \cite{MPII}. $\dagger$ represents the method reproduced on our platform without pre-trained backbone on the ImageNet dataset\cite{ImageNet}.}
    \begin{tabular}{l|cccccccc}
    \toprule
    \rowcolor{gray!30} Method & Head & Sho. & Elb. & Wri. & Hip & Knee & Ank. & Mean  \\
    \hline
    \multicolumn{9}{c}{Conventional Methods} \\
    \hline
    SBL-Res50\cite{SBL}  & 96.4 & 95.3 & 89.0 & 83.2 & 88.4 & 84.0 &  79.6 & 88.5 \\
    SBL-Res152\cite{SBL}  & 97.0 & 95.9 & 90.0 & 85.0 & 89.2 & 85.3 & 81.3 & 90.1 \\
    HRNet-W32$^\dagger$\cite{HRNet} & 96.5 & 94.7 & 89.0 & 83.8 & 86.7 & 82.9 & 79.4 & 88.0 \\
    TokenPose-T$^\dagger$\cite{TokenPose} & 91.9 & 87.3 & 75.3 & 66.6 & 78.4 & 71.1 & 65.4 & 77.5 \\
    TokenPose-L \cite{TokenPose} & \underline{97.1} & 95.9 & \underline{91.0} & 85.8 & 89.5 & 86.1 & 82.7 & 90.2 \\
    ViTPose-S$^\dagger$\cite{ViTPose} & 94.1 & 91.2 & 82.1 & 74.3 & 83.4 & 78.1 & 73.2 & 83.0 \\
    ViTPose-B$^\dagger$\cite{ViTPose} & 94.6 & 91.6 & 82.5 & 75.2 & 83.3 & 78.2 & 73.6 & 83.4 \\
    \hline
    \multicolumn{9}{c}{Lightweight Methods} \\
    \hline
    Lite-HRNet-18 \cite{Lite-HRNet} & - & - & - & - & - & - & - & 86.1 \\
    Lite-HRNet-30 \cite{Lite-HRNet} & - & - & - & - & - & - & - & 87.0 \\
    Dite-HRNet-18 \cite{Dite-HRNet} & - & - & - & - & - & - & - & 87.0 \\
    Dite-HRNet-18 \cite{Dite-HRNet} & - & - & - & - & - & - & - & 87.6 \\
    PPT-S\cite{PPT} & 96.6 & 94.9 & 87.6 & 81.3 & 87.1 & 82.4 & 76.7 & 87.3 \\
    PPT-B\cite{PPT} & 97.0 & 95.7 & 90.1 & 85.7 & 89.4 & 85.8 & 81.2 & 89.8 \\
    \hline 
    \textbf{MamKPD-S (Ours)} & \underline{97.1} & 96.0 & 90.6 & 86.0 & 89.8 & 86.7 & 82.9 & 90.3 \\
    \textbf{MamKPD-B (Ours)} & \underline{97.1} & \underline{96.1} & \underline{91.0} & \underline{86.7} & \underline{90.4} & \underline{87.1} & \underline{83.4} & \underline{90.7} \\
    \textbf{MamKPD-L (Ours)} & \textbf{97.5} & \textbf{96.4} & \textbf{91.6} & \textbf{87.7} & \textbf{90.7} & \textbf{87.8} & \textbf{84.9} & \textbf{91.3} \\
    \bottomrule
    \end{tabular}
    \label{tab:MPII}
\end{table}

\textbf{Comparisons with existing methods on AP-10K dataset.} We validate the effectiveness of MamKPD on the animal pose estimation task using the AP-10K dataset \cite{AP-10K}, with the experimental results shown in Table \ref{tab:AP-10K}. Compared to CNN-based and Transformer-based methods, our proposed MamKPD achieves competitive performance while being more lightweight. Specifically, compared to ViTPose-B, our method obtains a similar level of accuracy while saving about 92\% model parameters, demonstrating that our proposed MamKPD is also effective for 2D keypoint detection tasks in animals.

\begin{table}[htbp]
    \small
    \centering
    \setlength\tabcolsep{2pt}
    \caption{Comparisons with state-of-the-art methods on the AP-10K testing set\cite{AP-10K}. $\dagger$ represents the method reproduced on our platform without pre-trained backbone on the ImageNet dataset\cite{ImageNet}.}
    \begin{tabular}{l|ccccc|c}
    \toprule
    \rowcolor{gray!30} Method & $AP$    & $AP_{50}$ & $AP_{75}$ & $AP_{M}$    & $AP_{L}$ & \#Params  \\
    \hline
    SBL-Res50\cite{SBL}  & 68.1  & 92.3  & 74.0  & 51.0  & 68.8 & 34.0M \\
    SBL-Res101\cite{SBL}  & 68.1  & 92.2  & 74.2  & 53.4  & 68.8 & 53.0M \\

    HRNet-W32 \cite{HRNet} & 72.2 & 93.9 & 78.7 & 55.5 & 73.0 & 28.5M \\
    HRNet-W48 \cite{HRNet} & \underline{73.1} & \underline{93.7} & \underline{80.4}  & \underline{57.4} & \textbf{73.8} & 28.5M \\
    
    ViTPose-S \cite{ViTPose} & 68.7 & 93.0 & 75.1 &50.7 &69.1 & 24.3M \\
    ViTPose-B \cite{ViTPose} & \textbf{73.4} & \textbf{95.0} & \textbf{81.9} & \textbf{60.2} & \textbf{73.8} & 86.0M \\
    \hline
    \textbf{MamKPD-B (Ours)} & 71.2 & 92.7 & 78.2 & 55.5 & 71.9 & \textbf{7.1M} \\
   \textbf{MamKPD-L (Ours) }& 72.8 & 93.5 & 79.7 & 53.8 & 73.6 & \underline{12.4M} \\
    \bottomrule
    \end{tabular}
    \label{tab:AP-10K}
\end{table}

\subsubsection{Analysis of Inference Speed}
To evaluate the inference efficiency of MamKPD, we compare its model parameters, GFLOPs, and frame rate (FPS) with existing methods, as shown in Table \ref{tab:coco}.  MamKPD-L has the fewest parameters compared to conventional methods, and its inference speed surpasses conventional methods. Specifically, the inference speed of MamKPD-L is approximately 4 times faster than ViTPose-L's. It is worth noting that ViTPose was tested on an A100 GPU, while MamKPD-L was tested on a single 4090 GPU. Furthermore, compared to lightweight and real-time methods, MamKPD achieves the best results in both inference speed and localization accuracy. This demonstrates that the proposed MamKPD effectively balances performance and efficiency.

\subsection{Ablation Study}
We conduct a series of ablation studies on the MPII dataset \cite{MPII} to verify the structure's and critical elements' effectiveness. Unless specified, the ablation studies are conducted based on MamKPD-B.

\subsubsection{Ablation about MamKPD Architecture}
Table \ref{tab:ablation_structure} shows ablation studies about the number of blocks and the feature dimensions at each stage.

\textbf{Analysis of blocks.} First, we reduce the number of Mamba blocks in each stage of MamKPD, denoted as (a). The performance of (a) decreases by 1.1\%. Additionally, to investigate the impact of the network depth, we increase the number of stages in MamKPD to 4, denoted as (b). The performance of (b) decreases by 6.5\%, which may be due to the larger network size leading to overfitting.

\textbf{Analysis of dimension.} Methods (c) and (d) demonstrate the impact of the feature dimensions at each MamKPD stage on the model's performance. As shown in Table \ref{tab:ablation_structure}, the network's performance also declines as the feature dimensions decrease. This is expected, as the reduction in the number of learnable parameters leads to a weakened feature learning capability. However, while increasing the feature dimensions can improve the network's performance, it also substantially increases the number of parameters.

\begin{table}[htbp]
    \centering
    \setlength\tabcolsep{3pt}
    \footnotesize
    \caption{Ablation experiments about MamKPD architecture.}
    \begin{tabular}{c|cc|cc|c}
    \toprule
    \rowcolor{gray!30} Method & Blocks & Dimension & \#Param. & GFLOPs & Mean  \\
    \hline
    (a) & [1,2,3] & [96,192,384] & 4.0M & 2.9 & 89.6 \\
    (b) & [2,4,6,3] & [64,128,256,512] & 8.9M & 2.1 & 84.2 \\
    (c) & [2,4,6] & [64,128,256] & 3.2M & 2.0 & 84.4 \\
    (d) & [2,4,6] & [192,384,768] & 27.9M & 7.1 & 90.9 \\ 
    \hline
    Ours & [2,4,6] & [96,192,384] & 7.1M & 3.1 & 90.7 \\
    \bottomrule
    \end{tabular}
    \label{tab:ablation_structure}
\end{table}

\subsubsection{Ablation about Stem and CMM}

\begin{figure}
    \centering
    \includegraphics[scale=0.26]{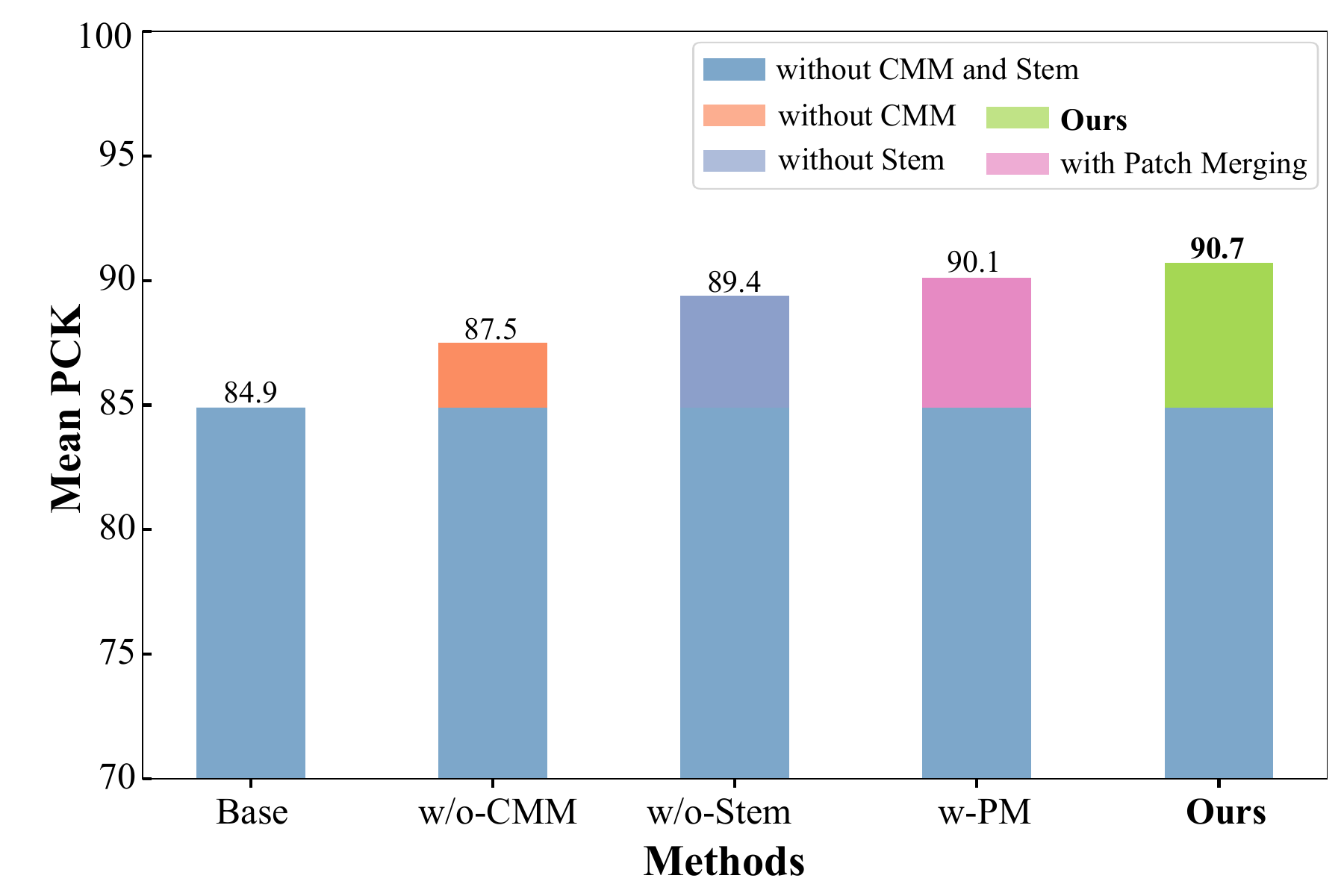}
    \caption{Ablation studies about Stem and CMM. ``Base'' represents the pure Mamba model. Different colors represent the performance improvements of other models relative to the baseline model.}
    \label{fig:ablation}
\end{figure}

In this section, we analyze the impact of two key components, Stem and CMM, on the performance of MamKPD. The experimental results are shown in Figure \ref{fig:ablation}. Here, the blue bars represent the baseline method (\textit{i.e.}, \textit{Base}), which does not incorporate the Stem and CMM modules. Different colors represent the performance improvements of other models relative to the baseline model.

\textbf{Analysis of Stem.} \textit{w/o-Stem} indicates the results of MamKPD without the Stem module. The performance of MamKPD drops by 1.3\% after removing the Stem. This is because, without the Stem, MamKPD lacks the initial noise filtering, which impacts its performance.

\textbf{Analysis of CMM.} Multi-scale contextual information is crucial for accurate keypoint localization as shown in Figure \ref{fig:ablation}, after removing the context modeling module (\textit{i.e.}, \textit{w/o-CMM}), the performance of MamKPD decreases by 3.2\%, due to the lack of inter-patch communication. Additionally, we replace the CMM with patch merging to validate the effectiveness of CMM, denoted as \textit{w-PM}. Patch merging helps the model extract multi-scale information, so \textit{w-PM}'s performance improves compared to w/o-CMM. However, patch merging struggles to capture dependencies between different patches, making its performance inferior to MamKPD's.

\begin{figure}
    \centering
    \includegraphics[width=\columnwidth]{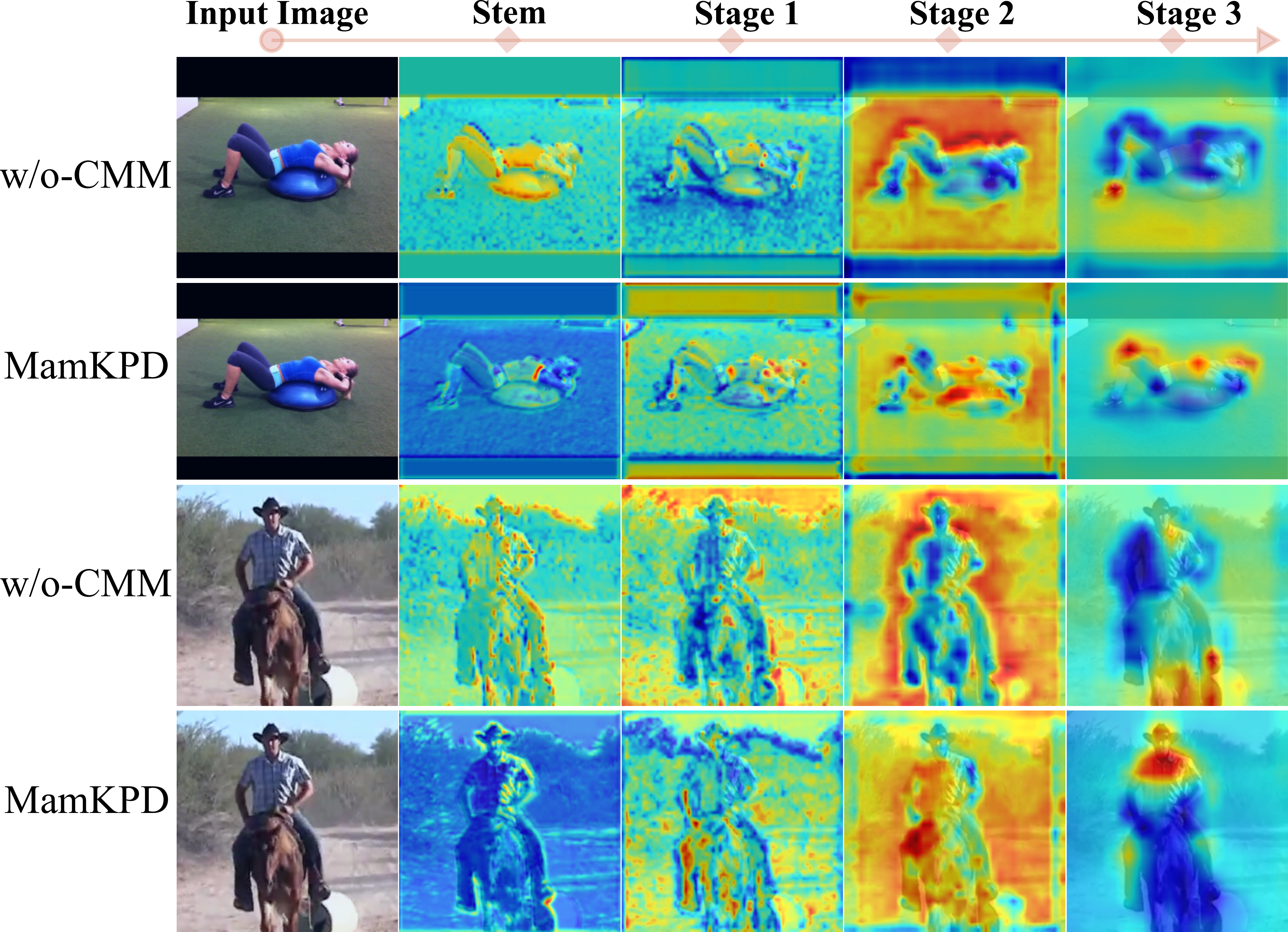}
    \caption{Ablation studies about CMM. The 1st and 3rd rows show the feature visualizations of MamKPD without the CMM. The 2nd and 4th rows display the feature visualizations of the proposed MamKPD.}
    \label{fig:cmm_feat}
\end{figure}

Additionally, to visually assess the impact of CMM on the model, we visualize the features at each stage, as shown in Figure \ref{fig:cmm_feat}. It can be observed that after incorporating CMM, the model can better focus on the regions where the instances are located.

\subsection{Qualitative Analysis}
\subsubsection{Feature Visualization}
\begin{figure}
    \centering
    \includegraphics[width=\columnwidth]{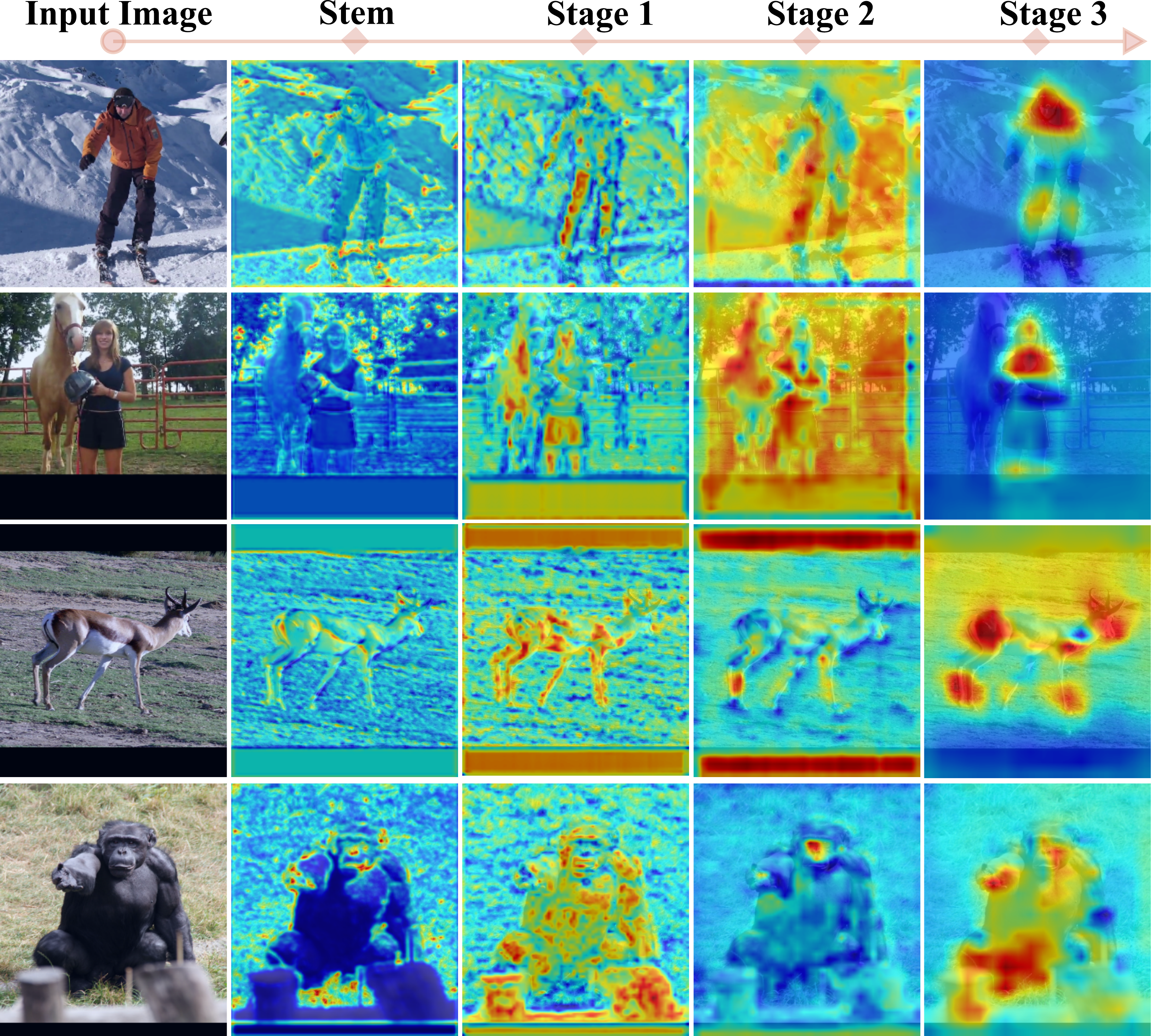}
    \caption{Feature visualization for each MamKPD's stage.}
    \label{fig:feature_vis}
\end{figure}

To illustrate how the proposed MamKPD effectively captures contextual information about poses, we visualize the output features of each stage in MamKPD, as shown in Figure \ref{fig:feature_vis}. We randomly choose samples from the MPII validation and AP-10K testing sets.

As we can see, the Stem effectively filters out irrelevant information while retaining structural information of the instance. Since Mamba is composed of linear layers with superior global modeling capabilities, the 1st and 2nd stages of MamKPD primarily focus on capturing the overall information of the input images. This aids the model in reasoning about accurate human poses based on foreground and background relationships. As shown in the feature map of the 4th stage, MamKPD focuses on the region where the instance is located.

\subsubsection{Pose Visualization}
\begin{figure}
    \centering
    \includegraphics[width=\columnwidth]{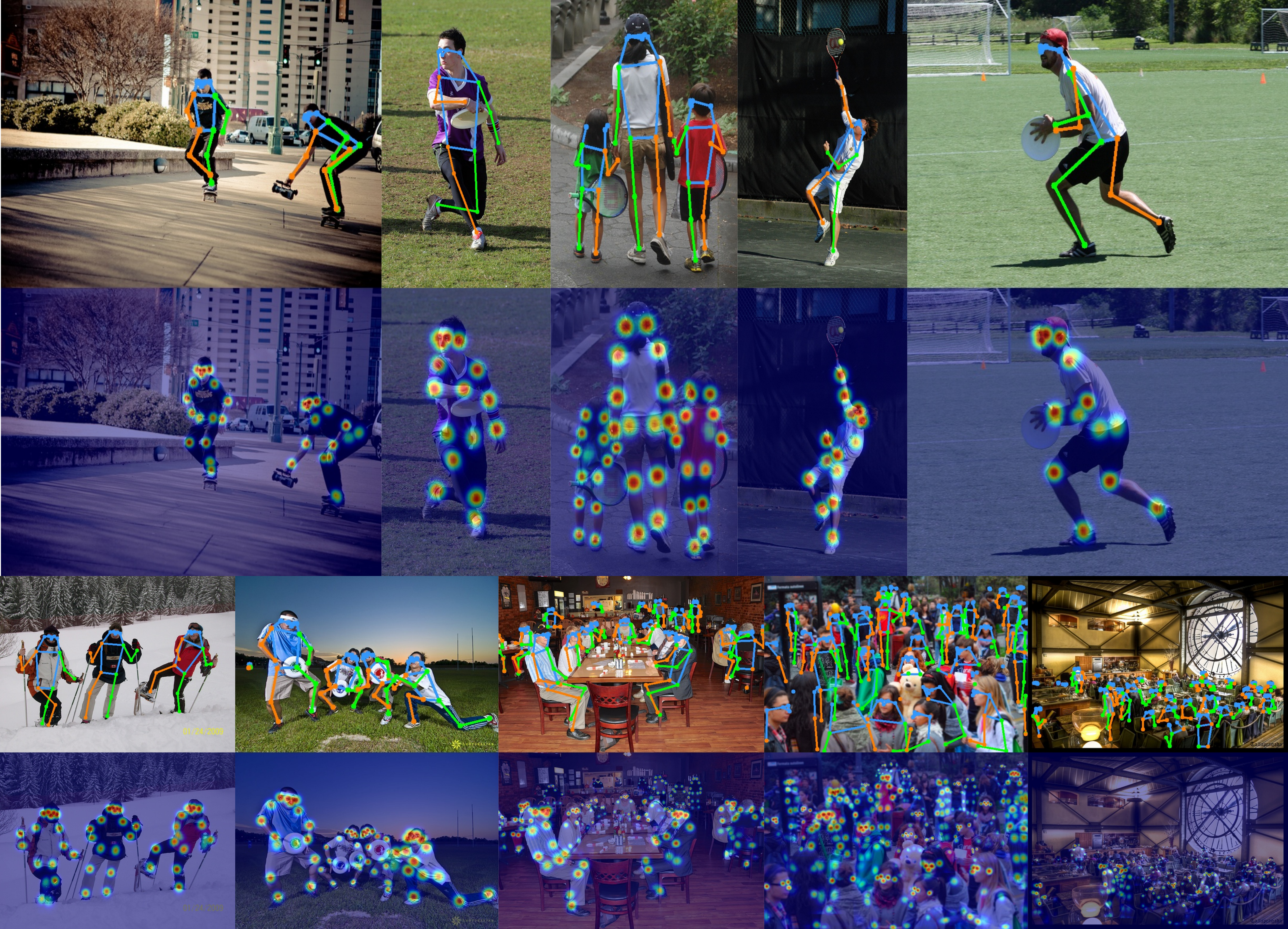}
    \caption{Visualization of the keypoint detection results of MamKPD on the COCO dataset \cite{COCO}.}
    \label{fig:coco_vis}
\end{figure}

\begin{figure}
    \centering
    \includegraphics[width=\columnwidth]{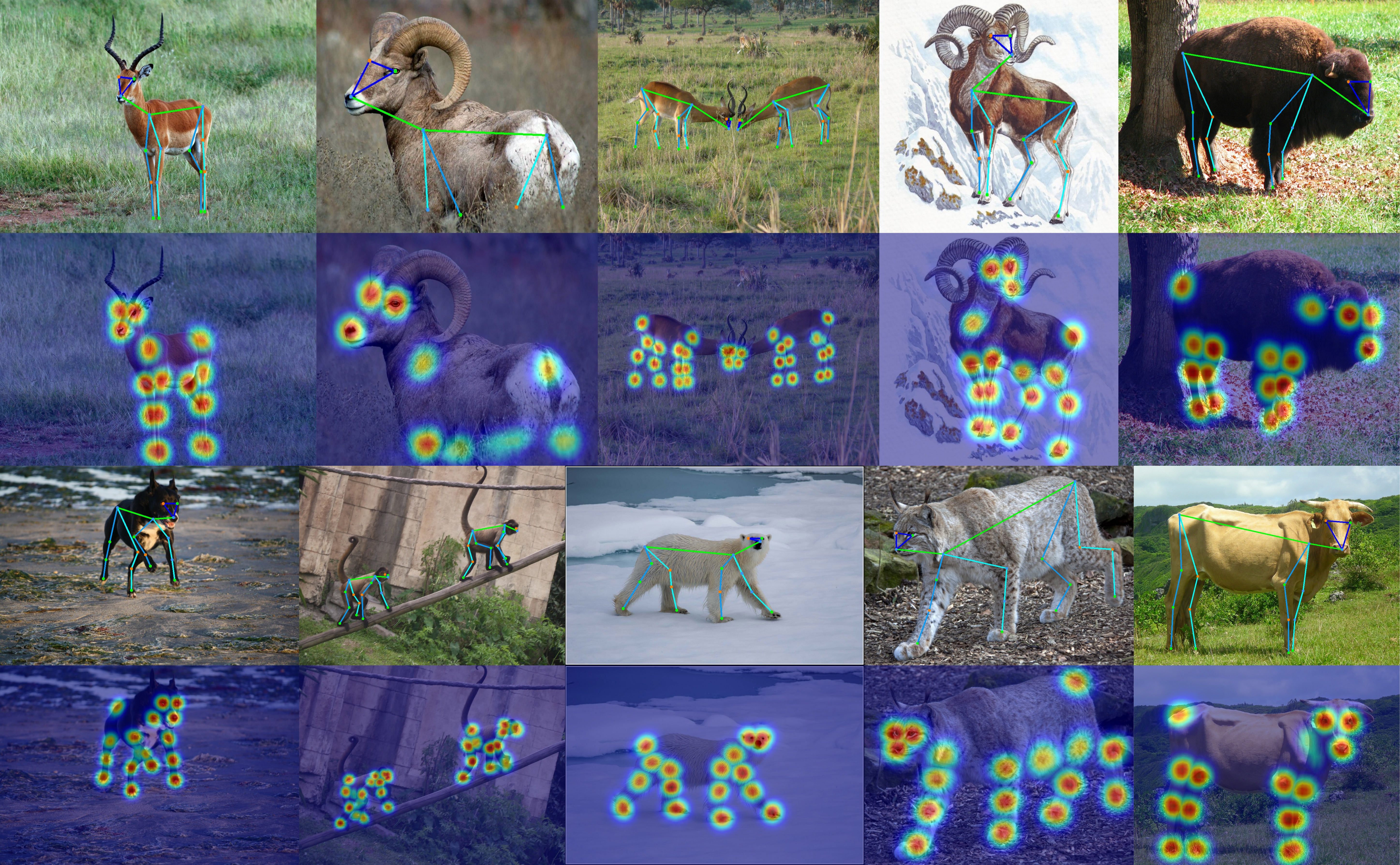}
    \caption{Visualization of the keypoint detection results of MamKPD on the AP-10K dataset \cite{AP-10K}.}
    \label{fig:ap10k_vis}
\end{figure}

We qualitatively present the keypoint detection results of MamKPD on the COCO (Figure \ref{fig:coco_vis}) and AP-10K (Figure \ref{fig:ap10k_vis}) datasets. As shown in Figures \ref{fig:coco_vis}, MamKPD accurately locates keypoints for each instance, whether in simple single-person scenes or crowded multi-person environments. Additionally, MamKPD demonstrates strong robustness to various human poses, such as running, bending, and handstands. Similarly, on the more complex outdoor datasets, MamKPD effectively locates keypoints for animals, as shown in Figure \ref{fig:ap10k_vis}. These experimental results highlight the effectiveness of the proposed MamKPD in 2D keypoint detection tasks.

\section{Conclusion}
This paper proposes a novel and efficient 2D keypoint detection framework, MamKPD, which is the first to validate the potential of Mamba in 2D keypoint detection tasks. We introduce a simple contextual modeling module to enhance the Mamba module's ability to model inter-patch dependencies. MamKPD achieves impressive results on both human and animal pose estimation datasets. Furthermore, MamKPD demonstrates superior inference efficiency, with MamKPD-L achieving a frame rate exceeding 1400 FPS on a single 4090 GPU, and MamKPD-B and MamKPD-S exceeding 2400 FPS. We hope that the proposed algorithm and its open-source implementation will meet some of the practical requirements in industrial applications.

\section{Acknowledgement}
This work was supported partly by the National Natural Science Foundation of China (Grant No. 62173045) and the China Postdoctoral Science Foundation under Grant Number 2024M750255.

{
    \small
    \bibliographystyle{ieeenat_fullname}
    \bibliography{main}

\begin{thebibliography}{62}
\providecommand{\natexlab}[1]{#1}
\providecommand{\url}[1]{\texttt{#1}}
\expandafter\ifx\csname urlstyle\endcsname\relax
  \providecommand{\doi}[1]{doi: #1}\else
  \providecommand{\doi}{doi: \begingroup \urlstyle{rm}\Url}\fi

\bibitem[Andriluka et~al.(2014)Andriluka, Pishchulin, Gehler, and Schiele]{MPII}
Mykhaylo Andriluka, Leonid Pishchulin, Peter~V. Gehler, and Bernt Schiele.
\newblock 2d human pose estimation: New benchmark and state of the art analysis.
\newblock In \emph{Proceedings of the {IEEE} Conference on Computer Vision and Pattern Recognition (CVPR)}, pages 3686--3693, 2014.

\bibitem[Cao et~al.(2021)Cao, Hidalgo, Simon, Wei, and Sheikh]{Openpose}
Zhe Cao, Gines Hidalgo, Tomas Simon, Shih{-}En Wei, and Yaser Sheikh.
\newblock Openpose: Realtime multi-person 2d pose estimation using part affinity fields.
\newblock \emph{{IEEE} Trans. Pattern Anal. Mach. Intell.}, 43\penalty0 (1):\penalty0 172--186, 2021.

\bibitem[Carreira et~al.(2016)Carreira, Agrawal, Fragkiadaki, and Malik]{IEF}
Joao Carreira, Pulkit Agrawal, Katerina Fragkiadaki, and Jitendra Malik.
\newblock Human pose estimation with iterative error feedback.
\newblock In \emph{Proceedings of the IEEE Conference on Computer Vision and Pattern Recognition (CVPR)}, pages 4733--4742, 2016.

\bibitem[Chen and Yuille(2014)]{chen2014articulated}
Xianjie Chen and Alan~L Yuille.
\newblock Articulated pose estimation by a graphical model with image dependent pairwise relations.
\newblock In \emph{Proceedings of the Neural Information Processing Systems (NeurIPS)}, 2014.

\bibitem[Chu et~al.(2017)Chu, Yang, Ouyang, Ma, Yuille, and Wang]{chu2017multi}
Xiao Chu, Wei Yang, Wanli Ouyang, Cheng Ma, Alan~L Yuille, and Xiaogang Wang.
\newblock Multi-context attention for human pose estimation.
\newblock In \emph{Proceedings of the IEEE Conference on Computer Vision and Pattern Recognition (CVPR)}, pages 1831--1840, 2017.

\bibitem[Dang et~al.(2022)Dang, Yin, and Zhang]{RPSTN}
Yonghao Dang, Jianqin Yin, and Shaojie Zhang.
\newblock Relation-based associative joint location for human pose estimation in videos.
\newblock \emph{{IEEE} Trans. Image Process.}, 31:\penalty0 3973--3986, 2022.

\bibitem[Dang et~al.(2024)Dang, Yin, Liu, Ding, Sun, and Hu]{DHRNet}
Yonghao Dang, Jianqin Yin, Liyuan Liu, Pengxiang Ding, Yuan Sun, and Yanzhu Hu.
\newblock Dhrnet: A dual-path hierarchical relation network for multi-person pose estimation.
\newblock \emph{Knowledge-Based Systems}, 300:\penalty0 112263, 2024.

\bibitem[Ding et~al.(2022)Ding, Deng, Zheng, Liu, Wang, Cheng, Bao, Chen, and Zeng]{I2RNet}
Yiwei Ding, Wenjin Deng, Yinglin Zheng, Pengfei Liu, Meihong Wang, Xuan Cheng, Jianmin Bao, Dong Chen, and Ming Zeng.
\newblock I{\({^2}\)}r-net: Intra- and inter-human relation network for multi-person pose estimation.
\newblock In \emph{Proceedings of the International Joint Conference on Artificial Intelligence (IJCAI)}, pages 855--862, 2022.

\bibitem[Fang et~al.(2022)Fang, Li, Tang, Xu, Zhu, Xiu, Li, and Lu]{AlphaPose}
H. Fang, J. Li, H. Tang, C. Xu, H. Zhu, Y. Xiu, Y. Li, and C. Lu.
\newblock Alphapose: Whole-body regional multi-person pose estimation and tracking in real-time.
\newblock \emph{{IEEE} Trans. Pattern Anal. Mach. Intell.}, \penalty0 (1):\penalty0 1--17, 2022.

\bibitem[Fu et~al.(2022)Fu, Dao, Saab, Thomas, Rudra, and R{\'e}]{fu2022hungry}
Daniel~Y Fu, Tri Dao, Khaled~K Saab, Armin~W Thomas, Atri Rudra, and Christopher R{\'e}.
\newblock Hungry hungry hippos: Towards language modeling with state space models.
\newblock \emph{arXiv preprint arXiv:2212.14052}, 2022.

\bibitem[Gu and Dao(2023)]{Mamba}
Albert Gu and Tri Dao.
\newblock Mamba: Linear-time sequence modeling with selective state spaces.
\newblock \emph{CoRR}, abs/2312.00752, 2023.

\bibitem[Gu et~al.(2020)Gu, Dao, Ermon, Rudra, and R{\'e}]{gu2020hippo}
Albert Gu, Tri Dao, Stefano Ermon, Atri Rudra, and Christopher R{\'e}.
\newblock Hippo: Recurrent memory with optimal polynomial projections.
\newblock In \emph{Proceedings of the Neural Information Processing Systems (NeurIPS)}, pages 1474--1487, 2020.

\bibitem[Gu et~al.(2021{\natexlab{a}})Gu, Goel, and R{\'e}]{gu2021efficiently}
Albert Gu, Karan Goel, and Christopher R{\'e}.
\newblock Efficiently modeling long sequences with structured state spaces.
\newblock \emph{arXiv preprint arXiv:2111.00396}, 2021{\natexlab{a}}.

\bibitem[Gu et~al.(2021{\natexlab{b}})Gu, Johnson, Goel, Saab, Dao, Rudra, and R{\'{e}}]{SSM}
Albert Gu, Isys Johnson, Karan Goel, Khaled Saab, Tri Dao, Atri Rudra, and Christopher R{\'{e}}.
\newblock Combining recurrent, convolutional, and continuous-time models with linear state space layers.
\newblock In \emph{Proceedings of the Neural Information Processing Systems (NeurIPS)}, pages 572--585, 2021{\natexlab{b}}.

\bibitem[Hu and Ramanan(2016)]{hu2016bottom}
Peiyun Hu and Deva Ramanan.
\newblock Bottom-up and top-down reasoning with hierarchical rectified gaussians.
\newblock In \emph{Proceedings of the IEEE Conference on Computer Vision and Pattern Recognition (CVPR)}, pages 5600--5609, 2016.

\bibitem[Hu et~al.(2023)Hu, Zheng, Zhou, Chen, and Sukthankar]{LAMP}
Shengnan Hu, Ce Zheng, Zixiang Zhou, Chen Chen, and Gita Sukthankar.
\newblock {LAMP:} leveraging language prompts for multi-person pose estimation.
\newblock In \emph{Proceedings of the {IEEE} International Conference on Intelligent Robots and Systems (IROS)}, pages 3759--3766, 2023.

\bibitem[Huang et~al.(2024{\natexlab{a}})Huang, Pei, You, Wang, Qian, and Xu]{huang2024localmamba}
Tao Huang, Xiaohuan Pei, Shan You, Fei Wang, Chen Qian, and Chang Xu.
\newblock Localmamba: Visual state space model with windowed selective scan.
\newblock \emph{arXiv preprint arXiv:2403.09338}, 2024{\natexlab{a}}.

\bibitem[Huang et~al.(2024{\natexlab{b}})Huang, Liu, Xian, and Qiu]{PoseMamba}
Yunlong Huang, Junshuo Liu, Ke Xian, and Robert~Caiming Qiu.
\newblock Posemamba: Monocular 3d human pose estimation with bidirectional global-local spatio-temporal state space model.
\newblock \emph{CoRR}, abs/2408.03540, 2024{\natexlab{b}}.

\bibitem[Jain et~al.(2013)Jain, Tompson, Andriluka, Taylor, and Bregler]{jain2013learning}
Arjun Jain, Jonathan Tompson, Mykhaylo Andriluka, Graham~W Taylor, and Christoph Bregler.
\newblock Learning human pose estimation features with convolutional networks.
\newblock \emph{arXiv preprint arXiv:1312.7302}, 2013.

\bibitem[Jiang et~al.(2023)Jiang, Lu, Zhang, Ma, Han, Lyu, Li, and Chen]{RTMPose}
Tao Jiang, Peng Lu, Li Zhang, Ningsheng Ma, Rui Han, Chengqi Lyu, Yining Li, and Kai Chen.
\newblock Rtmpose: Real-time multi-person pose estimation based on mmpose.
\newblock \emph{CoRR}, abs/2303.07399, 2023.

\bibitem[Jin et~al.(2022)Jin, Lee, and Lee]{OTPose}
Kyung{-}Min Jin, Gun{-}Hee Lee, and Seong{-}Whan Lee.
\newblock Otpose: Occlusion-aware transformer for pose estimation in sparsely-labeled videos.
\newblock In \emph{Proceedings of the {IEEE} International Conference on Systems, Man, and Cybernetics (SMC)}, pages 3255--3260, 2022.

\bibitem[Li et~al.(2021{\natexlab{a}})Li, Wang, Zhang, Xu, Xu, and Tu]{PRTR}
Ke Li, Shijie Wang, Xiang Zhang, Yifan Xu, Weijian Xu, and Zhuowen Tu.
\newblock Pose recognition with cascade transformers.
\newblock In \emph{Proceedings of the {IEEE/CVF} Conference on Computer Vision and Pattern Recognition (CVPR)}, pages 1944--1953, 2021{\natexlab{a}}.

\bibitem[Li et~al.(2024)Li, Li, Wang, He, Wang, Wang, and Qiao]{li2024videomamba}
Kunchang Li, Xinhao Li, Yi Wang, Yinan He, Yali Wang, Limin Wang, and Yu Qiao.
\newblock Videomamba: State space model for efficient video understanding.
\newblock \emph{arXiv preprint arXiv:2403.06977}, 2024.

\bibitem[Li et~al.(2022)Li, Zhang, Xiao, Zhang, and Bhanu]{Dite-HRNet}
Qun Li, Ziyi Zhang, Fu Xiao, Feng Zhang, and Bir Bhanu.
\newblock Dite-hrnet: Dynamic lightweight high-resolution network for human pose estimation.
\newblock In \emph{Proceedings of the International Joint Conference on Artificial Intelligence (IJCAI)}, pages 1095--1101, 2022.

\bibitem[Li et~al.(2021{\natexlab{b}})Li, Zhang, Wang, Yang, Yang, Xia, and Zhou]{TokenPose}
Yanjie Li, Shoukui Zhang, Zhicheng Wang, Sen Yang, Wankou Yang, Shu{-}Tao Xia, and Erjin Zhou.
\newblock Tokenpose: Learning keypoint tokens for human pose estimation.
\newblock In \emph{Proceedings of the IEEE/CVF International Conference on Computer Vision (ICCV)}, pages 11293--11302, 2021{\natexlab{b}}.

\bibitem[Lin et~al.(2014)Lin, Maire, Belongie, Hays, Perona, Ramanan, Doll{\'{a}}r, and Zitnick]{COCO}
Tsung{-}Yi Lin, Michael Maire, Serge~J. Belongie, James Hays, Pietro Perona, Deva Ramanan, Piotr Doll{\'{a}}r, and C.~Lawrence Zitnick.
\newblock Microsoft {COCO:} common objects in context.
\newblock In \emph{Proceedings of the European Conference on Computer Vision (ECCV)}, pages 740--755, 2014.

\bibitem[Liu et~al.(2023)Liu, Chen, Tan, Liu, Wang, Su, Li, Yao, Han, Ding, Zhao, and Wang]{GroupPose}
Huan Liu, Qiang Chen, Zichang Tan, Jiang{-}Jiang Liu, Jian Wang, Xiangbo Su, Xiaolong Li, Kun Yao, Junyu Han, Errui Ding, Yao Zhao, and Jingdong Wang.
\newblock Group pose: {A} simple baseline for end-to-end multi-person pose estimation.
\newblock In \emph{Proceedings of the {IEEE/CVF} International Conference on Computer Vision (ICCV)}, pages 14983--14992, 2023.

\bibitem[Liu et~al.(2024{\natexlab{a}})Liu, Yang, Zhou, Xi, Yu, Li, Liang, Shi, Yu, Zhang, Zheng, and Wang]{Swin-UMamba}
Jiarun Liu, Hao Yang, Hong{-}Yu Zhou, Yan Xi, Lequan Yu, Cheng Li, Yong Liang, Guangming Shi, Yizhou Yu, Shaoting Zhang, Hairong Zheng, and Shanshan Wang.
\newblock Swin-umamba: Mamba-based unet with imagenet-based pretraining.
\newblock In \emph{Proceedings of the Medical Image Computing and Computer Assisted Intervention (MICCAI)}, pages 615--625, 2024{\natexlab{a}}.

\bibitem[Liu et~al.(2024{\natexlab{b}})Liu, Tian, Zhao, Yu, Xie, Wang, Ye, and Liu]{Vmamba}
Yue Liu, Yunjie Tian, Yuzhong Zhao, Hongtian Yu, Lingxi Xie, Yaowei Wang, Qixiang Ye, and Yunfan Liu.
\newblock Vmamba: Visual state space model.
\newblock \emph{CoRR}, abs/2401.10166, 2024{\natexlab{b}}.

\bibitem[Loshchilov and Hutter(2019)]{Adam}
Ilya Loshchilov and Frank Hutter.
\newblock Decoupled weight decay regularization.
\newblock In \emph{Proceedings of the International Conference on Learning Representations (ICLR)}, 2019.

\bibitem[Lu et~al.(2024)Lu, Jiang, Li, Li, Chen, and Yang]{RTMO}
Peng Lu, Tao Jiang, Yining Li, Xiangtai Li, Kai Chen, and Wenming Yang.
\newblock {RTMO:} towards high-performance one-stage real-time multi-person pose estimation.
\newblock In \emph{Proceedings of the{IEEE/CVF} Conference on Computer Vision and Pattern Recognition (CVPR)}, pages 1491--1500, 2024.

\bibitem[Luvizon et~al.(2019)Luvizon, Tabia, and Picard]{luvizon2019human}
Diogo~C Luvizon, Hedi Tabia, and David Picard.
\newblock Human pose regression by combining indirect part detection and contextual information.
\newblock \emph{Computers \& Graphics}, 85:\penalty0 15--22, 2019.

\bibitem[Ma et~al.(2022)Ma, Wang, Chen, Kong, Chen, Liu, Yan, Tang, and Xie]{PPT}
Haoyu Ma, Zhe Wang, Yifei Chen, Deying Kong, Liangjian Chen, Xingwei Liu, Xiangyi Yan, Hao Tang, and Xiaohui Xie.
\newblock Ppt: Token-pruned pose transformer for monocular and multi-view human pose estimation.
\newblock In \emph{Proceedings of the European Conference on Computer Vision (ECCV)}, pages 424--442, 2022.

\bibitem[Mao et~al.(2021)Mao, Ge, Shen, Tian, Wang, and Wang]{TFPose}
Weian Mao, Yongtao Ge, Chunhua Shen, Zhi Tian, Xinlong Wang, and Zhibin Wang.
\newblock Tfpose: Direct human pose estimation with transformers.
\newblock \emph{CoRR}, 2021.

\bibitem[Mao et~al.(2022)Mao, Ge, Shen, Tian, Wang, Wang, and den Hengel]{mao2022poseur}
Weian Mao, Yongtao Ge, Chunhua Shen, Zhi Tian, Xinlong Wang, Zhibin Wang, and Anton~van den Hengel.
\newblock Poseur: Direct human pose regression with transformers.
\newblock In \emph{Proceedings of the European Conference on Computer Vision (ECCV)}, pages 72--88. Springer, 2022.

\bibitem[McNally et~al.(2022)McNally, Vats, Wong, and McPhee]{KAPAO}
William~J. McNally, Kanav Vats, Alexander Wong, and John McPhee.
\newblock Rethinking keypoint representations: Modeling keypoints and poses as objects for multi-person human pose estimation.
\newblock In \emph{Proceedings of the European Conference on Computer Vision (ECCV)}, pages 37--54, 2022.

\bibitem[Neff et~al.(2021)Neff, Sheth, Furgurson, Middleton, and Tabkhi]{EfficientHRNet}
Christopher Neff, Aneri Sheth, Steven Furgurson, John Middleton, and Hamed Tabkhi.
\newblock Efficienthrnet.
\newblock \emph{J. Real Time Image Process.}, 18\penalty0 (4):\penalty0 1037--1049, 2021.

\bibitem[Newell et~al.(2016)Newell, Yang, and Deng]{Hourglass}
Alejandro Newell, Kaiyu Yang, and Jia Deng.
\newblock Stacked hourglass networks for human pose estimation.
\newblock In \emph{Proceedings of the European Conference on Computer Vision (ECCV)}, pages 483--499, 2016.

\bibitem[Redmon et~al.(2016)Redmon, Divvala, Girshick, and Farhadi]{YOLO}
Joseph Redmon, Santosh~Kumar Divvala, Ross~B. Girshick, and Ali Farhadi.
\newblock You only look once: Unified, real-time object detection.
\newblock In \emph{Proceedings of the {IEEE} Conference on Computer Vision and Pattern Recognition (CVPR)}, pages 779--788, 2016.

\bibitem[Russakovsky et~al.(2015)Russakovsky, Deng, Su, Krause, Satheesh, Ma, Huang, Karpathy, Khosla, Bernstein, Berg, and Fei{-}Fei]{ImageNet}
Olga Russakovsky, Jia Deng, Hao Su, Jonathan Krause, Sanjeev Satheesh, Sean Ma, Zhiheng Huang, Andrej Karpathy, Aditya Khosla, Michael~S. Bernstein, Alexander~C. Berg, and Li Fei{-}Fei.
\newblock Imagenet large scale visual recognition challenge.
\newblock \emph{Int. J. Comput. Vis.}, 115\penalty0 (3):\penalty0 211--252, 2015.

\bibitem[Shi et~al.(2022)Shi, Wei, Li, Ren, and Tan]{PETR}
Dahu Shi, Xing Wei, Liangqi Li, Ye Ren, and Wenming Tan.
\newblock End-to-end multi-person pose estimation with transformers.
\newblock In \emph{Proceedings of the {IEEE/CVF} Conference on Computer Vision and Pattern Recognition (CVPR)}, pages 11069--11078, 2022.

\bibitem[Sun et~al.(2019)Sun, Xiao, Liu, and Wang]{HRNet}
Ke Sun, Bin Xiao, Dong Liu, and Jingdong Wang.
\newblock Deep high-resolution representation learning for human pose estimation.
\newblock In \emph{Proceedings of the {IEEE} Conference on Computer Vision and Pattern Recognition (CVPR)}, pages 5693--5703, 2019.

\bibitem[Sun et~al.(2017)Sun, Shang, Liang, and Wei]{sun2017compositional}
Xiao Sun, Jiaxiang Shang, Shuang Liang, and Yichen Wei.
\newblock Compositional human pose regression.
\newblock In \emph{Proceedings of the IEEE International Conference on Computer Vision (ICCV)}, pages 2602--2611, 2017.

\bibitem[Tan and Le(2019)]{EfficientNet}
Mingxing Tan and Quoc~V. Le.
\newblock Efficientnet: Rethinking model scaling for convolutional neural networks.
\newblock In \emph{Proceedings of the International Conference on Machine Learning (ICML)}, pages 6105--6114, 2019.

\bibitem[Tompson et~al.(2015)Tompson, Goroshin, Jain, LeCun, and Bregler]{tompson2015efficient}
Jonathan Tompson, Ross Goroshin, Arjun Jain, Yann LeCun, and Christoph Bregler.
\newblock Efficient object localization using convolutional networks.
\newblock In \emph{Proceedings of the IEEE Conference on Computer Vision and Pattern Recognition (CVPR)}, pages 648--656, 2015.

\bibitem[Tompson et~al.(2014)Tompson, Jain, LeCun, and Bregler]{tompson2014joint}
Jonathan~J Tompson, Arjun Jain, Yann LeCun, and Christoph Bregler.
\newblock Joint training of a convolutional network and a graphical model for human pose estimation.
\newblock In \emph{Proceedings of the Neural Information Processing Systems (NeurIPS)}, 2014.

\bibitem[Toshev and Szegedy(2014)]{Deeppose}
Alexander Toshev and Christian Szegedy.
\newblock Deeppose: Human pose estimation via deep neural networks.
\newblock In \emph{Proceedings of the IEEE Conference on Computer Vision and Pattern Recognition (CVPR)}, pages 1653--1660, 2014.

\bibitem[Wang et~al.(2024)Wang, Liu, Tang, Wu, Xu, Chou, and Wang]{GTPT}
Haonan Wang, Jie Liu, Jie Tang, Gangshan Wu, Bo Xu, Yanbing Chou, and Yong Wang.
\newblock Gtpt: Group-based token pruning transformer for efficient human pose estimation.
\newblock \emph{CoRR}, abs/2407.10756, 2024.

\bibitem[Wang et~al.(2022)Wang, Li, Cai, Chen, and Han]{Lite-Pose}
Yihan Wang, Muyang Li, Han Cai, Wei-Ming Chen, and Song Han.
\newblock Lite pose: Efficient architecture design for 2d human pose estimation.
\newblock In \emph{Proceedings of the IEEE/CVF Conference on Computer Vision and Pattern Recognition (CVPR)}, pages 13126--13136, 2022.

\bibitem[Xiao et~al.(2018)Xiao, Wu, and Wei]{SBL}
Bin Xiao, Haiping Wu, and Yichen Wei.
\newblock Simple baselines for human pose estimation and tracking.
\newblock In \emph{Proceedings of the European Conference on Computer Vision (ECCV)}, pages 472--487, 2018.

\bibitem[Xiong et~al.(2022)Xiong, Wang, Li, Luo, and Cao]{SwinPose}
Zinan Xiong, Chenxi Wang, Ying Li, Yan Luo, and Yu Cao.
\newblock Swin-pose: Swin transformer based human pose estimation.
\newblock In \emph{Proceedings of the IEEE International Conference on Multimedia Information Processing and Retrieval (MIPR)}, pages 228--233, 2022.

\bibitem[Xu et~al.(2022)Xu, Zhang, Zhang, and Tao]{ViTPose}
Yufei Xu, Jing Zhang, Qiming Zhang, and Dacheng Tao.
\newblock Vitpose: Simple vision transformer baselines for human pose estimation.
\newblock In \emph{Proceedings of the Advances in Neural Information Processing Systems (NeurIPS)}, 2022.

\bibitem[Yang et~al.(2024{\natexlab{a}})Yang, Chen, Espinosa, Ericsson, Wang, Liu, and Crowley]{yang2024plainmamba}
Chenhongyi Yang, Zehui Chen, Miguel Espinosa, Linus Ericsson, Zhenyu Wang, Jiaming Liu, and Elliot~J Crowley.
\newblock Plainmamba: Improving non-hierarchical mamba in visual recognition.
\newblock \emph{arXiv preprint arXiv:2403.17695}, 2024{\natexlab{a}}.

\bibitem[Yang et~al.(2017)Yang, Li, Ouyang, Li, and Wang]{yang2017learning}
Wei Yang, Shuang Li, Wanli Ouyang, Hongsheng Li, and Xiaogang Wang.
\newblock Learning feature pyramids for human pose estimation.
\newblock In \emph{Proceedings of the IEEE International Conference on Computer Vision (ICCV)}, pages 1281--1290, 2017.

\bibitem[Yang et~al.(2024{\natexlab{b}})Yang, Ma, Yao, Zhong, Zhang, and Wang]{yang2024remamber}
Yuhuan Yang, Chaofan Ma, Jiangchao Yao, Zhun Zhong, Ya Zhang, and Yanfeng Wang.
\newblock Remamber: Referring image segmentation with mamba twister.
\newblock \emph{arXiv preprint arXiv:2403.17839}, 2024{\natexlab{b}}.

\bibitem[Yu et~al.(2021{\natexlab{a}})Yu, Xiao, Gao, Yuan, Zhang, Sang, and Wang]{Lite-HRNet}
Changqian Yu, Bin Xiao, Changxin Gao, Lu Yuan, Lei Zhang, Nong Sang, and Jingdong Wang.
\newblock Lite-hrnet: {A} lightweight high-resolution network.
\newblock In \emph{Proceedings of the {IEEE} Conference on Computer Vision and Pattern Recognition (CVPR)}, pages 10440--10450, 2021{\natexlab{a}}.

\bibitem[Yu et~al.(2021{\natexlab{b}})Yu, Xu, Zhang, Zhao, Guan, and Tao]{AP-10K}
Hang Yu, Yufei Xu, Jing Zhang, Wei Zhao, Ziyu Guan, and Dacheng Tao.
\newblock {AP-10K:} {A} benchmark for animal pose estimation in the wild.
\newblock In \emph{Proceedings of the Neural Information Processing Systems (NeurIPS)}, 2021{\natexlab{b}}.

\bibitem[Zhang et~al.(2024)Zhang, Zhu, Wang, Zhang, Chen, and Ye]{MambaSurvey}
Hanwei Zhang, Ying Zhu, Dan Wang, Lijun Zhang, Tianxiang Chen, and Zi Ye.
\newblock A survey on visual mamba.
\newblock \emph{CoRR}, abs/2404.15956, 2024.

\bibitem[Zhang et~al.(2021)Zhang, Fang, Wang, and Liu]{EfficientPose}
Wenqiang Zhang, Jiemin Fang, Xinggang Wang, and Wenyu Liu.
\newblock Efficientpose: Efficient human pose estimation with neural architecture search.
\newblock \emph{Comput. Vis. Media}, 7\penalty0 (3):\penalty0 335--347, 2021.

\bibitem[Zhang et~al.(2023)Zhang, Wang, Chen, Xu, Zhang, and Tao]{CLAMP}
Xu Zhang, Wen Wang, Zhe Chen, Yufei Xu, Jing Zhang, and Dacheng Tao.
\newblock {CLAMP:} prompt-based contrastive learning for connecting language and animal pose.
\newblock In \emph{Proceedings of the {IEEE} Conference on Computer Vision and Pattern Recognition (CVPR)}, pages 23272--23281, 2023.

\bibitem[Zheng et~al.(2024)Zheng, Wu, Chen, Yang, Zhu, Shen, Kehtarnavaz, and Shah]{VR}
Ce Zheng, Wenhan Wu, Chen Chen, Taojiannan Yang, Sijie Zhu, Ju Shen, Nasser Kehtarnavaz, and Mubarak Shah.
\newblock Deep learning-based human pose estimation: {A} survey.
\newblock \emph{{ACM} Comput. Surv.}, 56\penalty0 (1):\penalty0 11:1--11:37, 2024.

\bibitem[Zhu et~al.(2024)Zhu, Liao, Zhang, Wang, Liu, and Wang]{Vim}
Lianghui Zhu, Bencheng Liao, Qian Zhang, Xinlong Wang, Wenyu Liu, and Xinggang Wang.
\newblock Vision mamba: Efficient visual representation learning with bidirectional state space model.
\newblock In \emph{Proceedings of the International Conference on Machine Learning (ICML)}, 2024.

\end{thebibliography}
}


\end{document}